\pdfoutput=1

\documentclass[11pt]{article}

\usepackage[]{emnlp2021}

\usepackage{times}
\usepackage{latexsym}

\usepackage{amsmath}

\usepackage[T1]{fontenc}

\usepackage[utf8]{inputenc}

\usepackage{microtype}

\usepackage{graphicx}
\usepackage{subcaption}

\usepackage{multirow}
\newif\ifshowedits
\showeditsfalse

\usepackage{xcolor}
\usepackage{soul}
\definecolor{customorangecolor}{HTML}{ffc46b}
\colorlet{customorangecolor}{customorangecolor!90}
\DeclareRobustCommand{\hlorange}[1]{{\sethlcolor{customorangecolor}\hl{#1}}}
\definecolor{customgreencolor}{HTML}{b9e0a5}
\colorlet{customgreencolor}{customgreencolor!90}
\DeclareRobustCommand{\hlgreen}[1]{{\sethlcolor{customgreencolor}\hl{#1}}}

\title{NUS-IDS at FinCausal 2021: Dependency Tree in Graph Neural Network for Better Cause-Effect Span Detection}

\author{Fiona Anting Tan, See-Kiong Ng \\
 Institute of Data Science \\
National University of Singapore, Singapore \\
  \texttt{tan.f@u.nus.edu, seekiong@nus.edu.sg} }

\date{}

\begin{document}
\maketitle
\begin{abstract}
Automatic identification of cause-effect spans in financial documents is important for causality modelling and understanding reasons that lead to financial events. To exploit the observation that words are more connected to other words with the same cause-effect type in a dependency tree, we construct useful graph embeddings by incorporating dependency relation features through a graph neural network. Our model builds on a baseline BERT token classifier with Viterbi decoding, and outperforms this baseline in cross-validation and during the competition. In the official run of FinCausal 2021, we obtained Precision, Recall, and F1 scores of $95.56\%$, $95.56\%$ and $95.57\%$ that all ranked 1\textsuperscript{st} place, and an Exact Match score of $86.05\%$ which ranked 3\textsuperscript{rd} place.
\end{abstract}

\section{Introduction}
\label{sec:introduction}
We worked on the shared task of FinCausal 2021 \citep{Mariko-fincausal-2021} that aims to identify cause and effect spans in financial news. This task builds on the previous shared Task 2 \citep{mariko-etal-2020-financial} by introducing additional annotated data.

\paragraph{Contributions:} We propose a solution to include dependency relations in a sentence to improve identification of cause-effect spans. We do so by representing dependency relations in a graph neural network.
Our model is an extension of a baseline BERT token classifier with Viterbi decoding \citep{kao-etal-2020-ntunlpl}, and outperforms this baseline in cross-validation and test settings. This improvement also holds for two BERT pretrained language models that were experimented with. During the competition, we ranked 1\textsuperscript{st} for Precision, Recall, and F1 scores and  3\textsuperscript{rd} for Exact Match score.

\paragraph{Organization:} Section \ref{sec:method} outlines our approach for this task. Section \ref{sec:data} introduces the task dataset and evaluation datasets. Our results are presented and discussed in Section \ref{sec:results} while Section \ref{sec:conclusion} concludes with some future directions.

\section{Our Approach} 
\label{sec:method}
In this section, we outline our approach \footnote{Our source code is available at \url{https://github.com/tanfiona/CauseEffectDetection}.}. Additional architectural and experimental details are provided in the \hyperref[sec:appendix]{Appendix}.

\subsection{Framing the Task} 
\label{sec:framing}
\begin{figure}[htpb]
\centering
\includegraphics[width=1\linewidth]{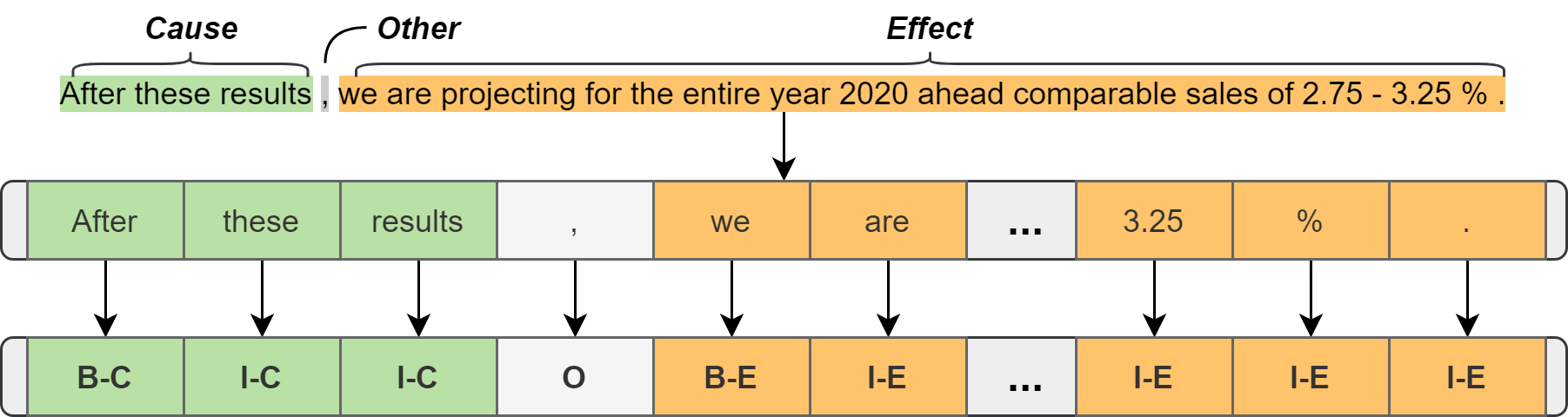}
\caption{Illustrative training example (ID: 0477.00020) with Cause (C) and Effect (E) spans highlighted in green and orange respectively. We include Begin (B), Inside (I) and Outside (O) prefixes to create 5 labels in our token classification task.}
\label{fig:example}
\end{figure}

Given an example document, which could be one or multiple sentence(s) long, the task is to identify the cause and effect substrings. We converted the span detection task into a token classification task, similar to many state-of-the-art methodologies for span detection \citep{pavlopoulos-etal-2021-semeval} and Named Entity Recognition \citep{lample-etal-2016-neural, tan2020boundary} tasks. Figure \ref{fig:example} demonstrates an example sentence that has its Cause (C) span highlighted in green, and Effect (E) span highlighted in orange, while all other spans are highlighted in grey. The sentence was tokenized and subsequently aligned against the target token labels. We included the BIO format (Begin, Inside, Outside) \citep{ramshaw-marcus-1995-text} in our labels to better identify the start of spans. Thus, we have five labels: B-C, I-C, B-E, I-E and O.

\subsection{Baseline}
\label{sec:baseline}
\citeauthor{kao-etal-2020-ntunlpl} demonstrated that their BIO tagging scheme with a Viterbi decoder \citep{1054010} that utilised BERT-encoded document representations is useful for this cause-effect span detection task. Their model topped the competition last year across all metrics. Therefore, we adapted their pipeline and proposed distinct additions highlighted later in Section \ref{sec:dependency} for improved performance. In the immediate subsections, we motivate the benefits in retaining the key components of \citeauthor{kao-etal-2020-ntunlpl}'s model, and highlight any differences in our approach.

\subsubsection{BERT Embeddings}
We employed the Bidirectional Encoder Representations from Transformers (BERT) \citep{devlin-etal-2019-bert} for its tokenizer and encoder model, fine-tuned on our task. We used pretrained language models from Huggingface \citep{wolf-etal-2020-transformers}. Apart from \texttt{bert-base-cased}, we also used \texttt{bert-large-cased} for improved performance.

\subsubsection{Viterbi Decoder}
The Viterbi decoding algorithm is only applied during evaluation. Since the true cause-effect spans are consecutive sequences, but the token classifications from the neural network could produce non-consecutive sequences, the Viterbi algorithm serves as a forward error correction technique and is an important element for the success of this pipeline.

\subsubsection{Parts-of-Speech}
\citet{kao-etal-2020-ntunlpl} showed that Parts-of-Speech (POS) did not improve their model performance. We reconfirm this finding later in Section \ref{ssec:components}. However, we found POS features to be useful inclusions for our proposed model.

\begin{figure*}[htbp]
    \centering
    \begin{subfigure}[b]{0.32\textwidth}
        \includegraphics[width=1\linewidth]{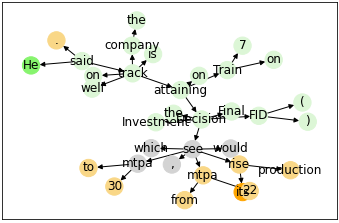}
    \end{subfigure}
    \begin{subfigure}[b]{0.32\textwidth}
        \includegraphics[width=1\linewidth]{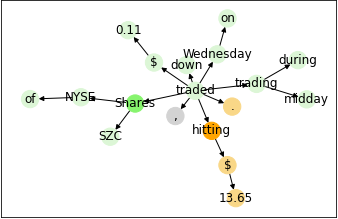}
    \end{subfigure}
        \begin{subfigure}[b]{0.32\textwidth}
        \includegraphics[width=1\linewidth]{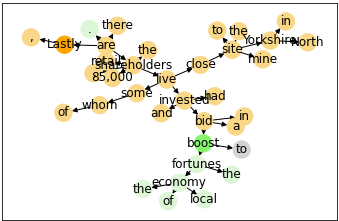}
    \end{subfigure}
    \caption{Dependency-tree represented as directed graphs, with Cause, Effect, and Other spans highlighted in green, orange and grey respectively.}
    \label{fig:depgraph}
\end{figure*}

\subsection{Dependency Tree}
\label{sec:dependency}
Our key contribution is the inclusion of dependency tree relations into the neural network for improved cause-effect token classification. Dependencies in text can be mapped into a directed graph representation, where nodes represent words and edges represent the dependency relation. Figure \ref{fig:depgraph} shows some example sentences, highlighted by their Cause and Effect labels. We notice that there is a tendency for words of the same cause-effect label to be more connected in these graphs and thus wish to incorporate these information into the model as features.

\begin{figure}[htpb]
    \centering
    \includegraphics[width=1\linewidth]{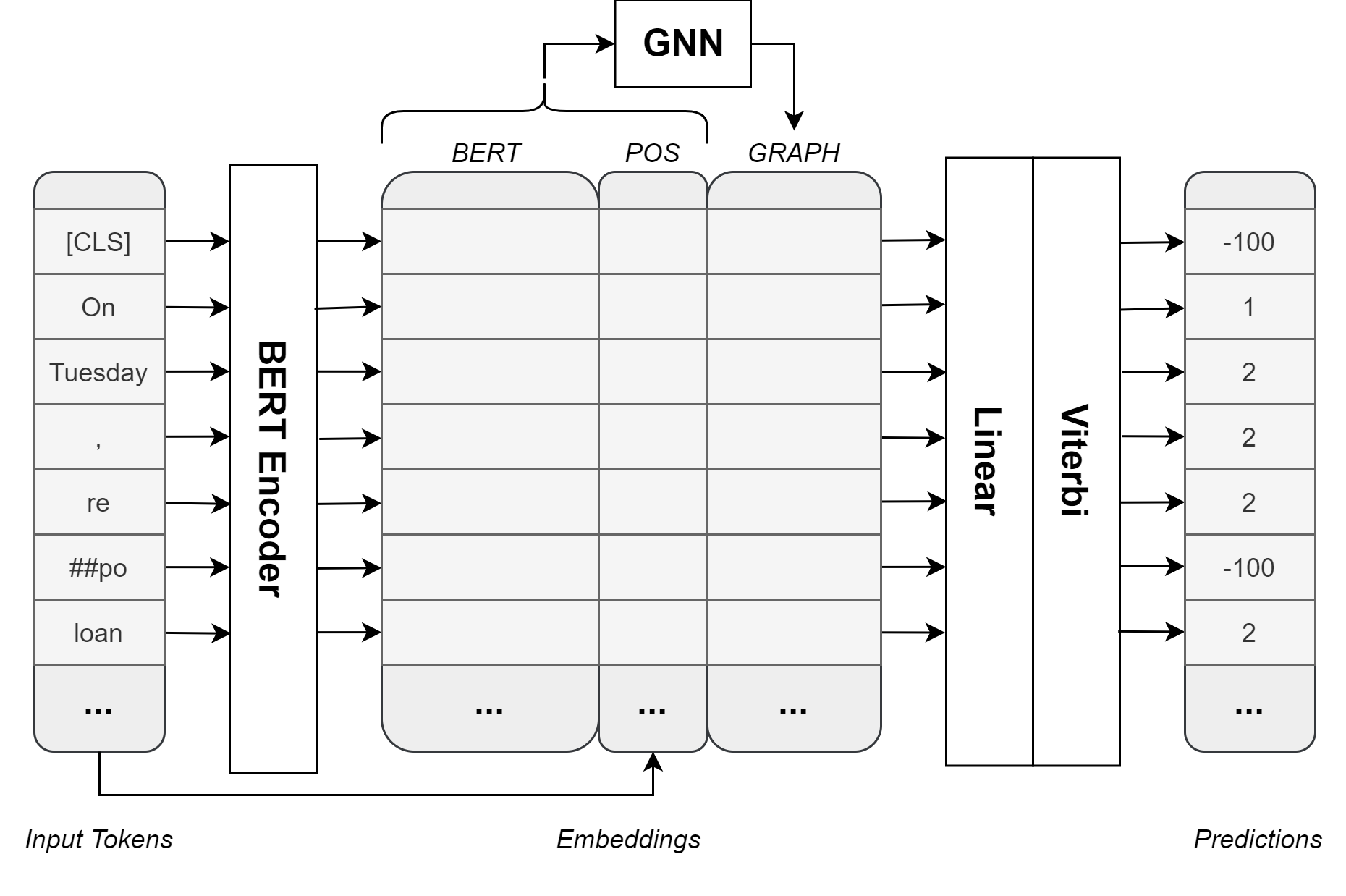}
    \caption{Neural network pipeline. Appendix \ref{ssec:architecture} outlines this further.}
    \label{fig:neuralnetwork}
\end{figure}

Figure \ref{fig:neuralnetwork} reflects our neural network model, with the addition of our GNN module that produces graph representations, which are then concatenated with the BERT and POS embeddings and fed into a linear layer for token classification. The following subsections describe the GNN module further.

\begin{table*}[]
\centering
\begin{tabular}{lllcccc}\hline
\multicolumn{2}{c}{Model} & & Precision & Recall & F1 & ExactMatch \\\hline
\multicolumn{7}{l}{\textbf{A. \texttt{bert-base-cased}}} \\
1 & Baseline & & 95.62 & 95.90 & 95.76 & 88.18 \\\hline
2 & \hspace*{0.3cm}+ Node features, BiLSTM & & 95.42 & 95.95 & 95.68 & 88.07 \\
3 & Baseline w/ POS & & 95.39\textasciicircum{} & 96.02 & 95.70 & 88.24 \\
4 & \hspace*{0.3cm}+ Node features & & 95.66 & 95.81 & 95.73 & 88.20 \\
5 & \hspace*{0.3cm}+ BiLSTM & & 95.46 & 95.99 & 95.72 & 88.06 \\
6 & \hspace*{0.3cm}+ Node features, BiLSTM (Proposed) & & \textbf{95.73} & \textbf{96.05\textasciicircum{}} & \textbf{95.89} & \textbf{88.35} \\
 &  &  &  \\
\multicolumn{7}{l}{\textbf{B. \texttt{bert-large-cased}}} \\
7 & Baseline & & 95.64 & 96.01 & 95.82 & 87.65 \\\hline
8 & \hspace*{0.3cm}+ Node features, BiLSTM & & 93.75* & 95.56 & 94.61* & 86.44 \\
9 & Baseline w/ POS & & 94.46 & 96.08 & 95.22 & 87.28 \\
10 & \hspace*{0.3cm}+ Node features & & 95.61 & 95.98 & 95.79 & 88.24 \\
11 & \hspace*{0.3cm}+ BiLSTM & & 95.60 & 95.78\textasciicircum{} & 95.69 & 87.94 \\
12 & \hspace*{0.3cm}+ Node features, BiLSTM (Proposed) & & \textbf{95.72} & \textbf{96.11} & \textbf{95.91} & \textbf{88.35}\\\hline
\end{tabular}
\caption{Average evaluation results over cross-validation sets from 5 random seeds, each with 3 folds. \emph{Notes.} Scores are reported in percentages (\%). Best score per Panel per column is bolded. Baseline models are our replications of the models introduced by \citet{kao-etal-2020-ntunlpl}. For each Panel A and B, Paired T-test of the models was conducted against Row 1 and 7 respectively, with statistical significance indicated by: ***$<0.05$, **$<0.10$, *$<0.15$, \textasciicircum$<0.20$.}
\label{tab:results}
\end{table*}

\begin{table*}[]
\centering
\begin{tabular}{llclclclc}
\hline
\multicolumn{1}{c}{Model} & \multicolumn{1}{l}{\textbf{}} & \multicolumn{1}{c}{Precision} & \multicolumn{1}{l}{\textbf{}} & \multicolumn{1}{c}{Recall} & \multicolumn{1}{l}{\textbf{}} & \multicolumn{1}{c}{F1} & \multicolumn{1}{l}{\textbf{}} & \multicolumn{1}{c}{ExactMatch} \\ \hline
Baseline &  & 93.47 &  & 93.42 &  & 93.65 &  & 80.25 \\
Proposed (\texttt{bert-base-cased}) &  & 94.24 &  & 94.21 &  & 94.37 &  & 83.23 \\
Proposed (\texttt{bert-large-cased}) & \textbf{} & \textbf{95.56} & \textbf{} & \textbf{95.56} & \textbf{} & \textbf{95.57} & \textbf{} & 86.05 \\
Best Score & \textbf{} & \textbf{95.56} & \textbf{} & \textbf{95.56} & \textbf{} & \textbf{95.57} & \textbf{} & \textbf{87.77} \\
 &  & \multicolumn{1}{l}{} &  & \multicolumn{1}{l}{} &  & \multicolumn{1}{l}{} &  & \multicolumn{1}{l}{} \\
\textit{Our Ranking} &  & \textit{1\textsuperscript{st}} & \textit{} & \textit{1\textsuperscript{st}} & \textit{} & \textit{1\textsuperscript{st}} & \textit{} & \textit{3\textsuperscript{rd}}\\\hline
\end{tabular}
\caption{Results over test sets submitted to Codalab (As of 01 September 2021). \emph{Notes.} Scores are reported in percentages (\%). Best score per column is bolded. The models of the first three rows corresponds to Rows 1, 6, and 12 in Table \ref{tab:results} for CV respectively.}
\label{tab:testing}
\end{table*}

\subsubsection{Document to Graph}
\label{ssec:doc2graph}
Each example was represented as a directed graph, where nodes are token features (the concatenation of BERT and POS embeddings), while edges are directed connections pointing head to tail tokens \footnote{If head or tail words are split into multiple tokens, each head (sub) piece is connected to each tail (sub) piece.} based on dependency tree parsing \footnote{Stanza \citep{qi-etal-2020-stanza} was used for dependency parsing.}.

\subsubsection{Graph Neural Network}
\label{ssec:gnn}
Our graph neural network (GNN) comprised of two graph convolutional layers for message passing across dependency relations that are two steps apart. 
Specifically, we used SAGEConv operator \citep{DBLP:conf/nips/HamiltonYL17} for its ability to include node features and generate embeddings by aggregating a node's neighbouring information. To capture the long-term contextual dependencies in both forward and backwards order of the original sentence, we added a bi-directional long short-term memory (BiLSTM) layer \citep{10.1162/neco.1997.9.8.1735} onto the graph embeddings. 

\section{Data}
\label{sec:data}

\subsection{Task Dataset}
\label{ssec:dataset}
Our main dataset is the FinCausal 2021 dataset \citep{Mariko-fincausal-2021} comprising of $2393$ train and $638$ competition test examples.

\subsection{Evaluation}
\label{ssec:evaluation}
We evaluated our proposed models in cross-validation (CV) against the Viterbi BERT model \citep{kao-etal-2020-ntunlpl} that achieved first place in the previous run of this shared task. We refer to this model as the Baseline in subsequent sections. To check if our proposed models has statistically significant improvements from the Baseline, we adapted \citet{DBLP:journals/neco/Dietterich98}'s approach using a 3-fold CV setup for 5 iterations, each iteration initialised with a random seed \footnote{The 5 random seeds used were $916, 703, 443, 229, 585$}. This gives us $5*3=15$ sets of evaluation results to perform Paired T-Tests for statistical significance.

To obtain test predictions for submission on Codalab \footnote{\url{https://competitions.codalab.org/competitions/33102}}, we used a new $seed=123$ to train on the full train data and applied the model onto the unseen competition test data for online submission.

\section{Results and Analysis}
\label{sec:results}
Table \ref{tab:results} reflects the model performances in CV, while Table \ref{tab:testing} reflects the results during the competition. In both cases, we demonstrate that our model (Proposed) surpasses the Baseline.

For the CV setting, Proposed (Row 6) obtained $95.73\%$ Precision (P), $96.05\%$ Recall (R), $95.89\%$ F1 and $88.35\%$ exact match (EM) scores. These exceed the Baseline (Row 1) by $0.11\%$, $0.15\%$, $0.13\%$ and $0.18\%$ respectively \footnote{We did not obtain P-values ($<5\%$) of statistical significance when comparing Proposed against Baseline.}. For test setting, large performance increments were observed: Proposed achieved P/R/F1/EM scores of $94.24\%$, $94.21\%$, $94.37\%$ and $83.23\%$, which are improvements from Baseline by a magnitude of $0.77\%$, $0.78\%$, $0.72\%$ and $2.98\%$ respectively \footnote{We were unable to run repeated iterations to conduct Paired T-Tests for significance testing in competition test sets as we do not have access to the true labels.}.

\subsection{Size of Pretrained Models}
The two sections of Table \ref{tab:results} shows that the inclusion of dependency-based graph embeddings improved performance against Baseline irregardless of the pretrained BERT model choice. 

Between the two investigated BERT models, \texttt{bert-large-cased} outperforms \texttt{bert-base-cased} by a small amount in CV and by a significant amount in testing across metrics. With \texttt{bert-large-cased}, the Proposed model achieved P/R/F1/EM scores of $95.56\%$, $95.56\%$, $95.57\%$ and $86.05\%$ during the competition, which are improvements from Baseline by a magnitude of $2.09\%$, $2.14\%$, $1.92\%$ and $5.80\%$ respectively \footnote{We did not upload a Baseline model using \texttt{bert-large-cased}, which would have allowed for a fairer comparison, due to time and upload constraints.}.

\subsection{Features and Layers}
\label{ssec:components}

Table \ref{tab:results} also includes results from CV experiments where we removed components of our model. In this subsection, we discuss the importance of each component in the context of \texttt{bert-base-cased}, but note that similar findings persisted in the \texttt{bert-large-cased}.

\paragraph{POS:} Inclusion of the POS into the Baseline led to mixed outcomes across the four metrics against the Baseline (Row 3 vs Row 1). However, adding POS features in Proposed improved performance in all metrics (Row 6 vs Row 2).

\paragraph{Node features:} We reran a model that takes in nodes with no features (i.e. all nodes are represented by ``$1$''). The results from this model corresponds to Row 5. No obvious improvements from Baseline (Row 1) was found, however, the performance was consistently worse off than Proposed for all metrics (Row 6), suggesting that informative node features are important to include in the GNN.

\paragraph{BiLSTM:} Comparing our proposed model with (Row 6) and without (Row 4) the BiLSTM layer suggested the layer was an important addition. Our hypothesis is that the BiLSTM helps to align the graph embeddings into a sequential manner corresponding to the original token order. A simple example is the punctuation full-stop ``.'' in Figure \ref{fig:depgraph}. The target label of the full-stop in these cases coincides with the immediate label of the word before it. However, our dependency tree attributed the full-stop as a tail of another word far away from it in the sentence (E.g. \textit{``said''} $\rightarrow$ {``.''}).



\section{Conclusions and Future Work}
\label{sec:conclusion}
We have demonstrated the benefits of including dependency tree features as graph embeddings in a neural network model for better cause-effect span detection. A key caveat of our approach, which requires further research, is that dependency parsing occurs within sentences, resulting in disconnected graphs for examples with multiple sentences \footnote{Preliminary experiments to tie coreferential entities together to link dependency across sentences did not produce fruitful results.}. Another future work is to apply our cause-effect detection model trained on financial texts onto other domains (E.g. academic journals) to study its generalizability.




\bibliographystyle{acl_natbib}
\bibliography{bibs/anthology, bibs/acl2021, bibs/publications}

\clearpage
\newpage

\appendix
\section{Appendix}
\label{sec:appendix}

\subsection{Architecture of Proposed Model}
\label{ssec:architecture}
An example with $n$ tokens can be represented by the vector of tokens $\underline{w} = (w_0, w_1, ..., w_n)^{\mathsf{T}}$, where $w$ refers to a BERT input token. A single token in this vector is represented by $w_i$, where $i$ denotes the location index of the token within the example. The tokenized example also has a matrix of POS features $V = (\underline{v_0}, \underline{v_1}, ..., \underline{v_n})^{\mathsf{T}}$, where each $\underline{v_i} \in {\rm I\!R}^{d_{POS}}$ refers to a one-hot encoding across all POS tags, and so $d_{POS}$ reflects the number of possible POS tags.

To obtain BERT representations ($r$), we run the tokenized sequence vector through the BERT encoder ($T$) to obtain $r$ for each token $i$. Thus, we have that,
\begin{align}
\underline{{r}_i} = T_w(\underline{w})_i, && \underline{{r}_i} \in {\rm I\!R}^{d_{BERT}}
\end{align}
where $d_{BERT}$ refers to the output dimension for BERT encoder.

A dropout layer is represented by $\delta \sim Bernoulli(\rho)$, where $\rho$ refers to the dropout probability. We apply the dropout layer onto the BERT representations as follows,
\begin{align}
\tilde{\underline{{r}_i}} = \delta * \underline{{r}_i}, && {r}_i \in {\rm I\!R}^{d_{BERT}}
\end{align}

We combine the BERT and POS representations of each token together by a simple concatenation along the feature dimension.
\begin{align}
\underline{{r_2}_i} = [\tilde{\underline{{r}_i}}, \underline{v_i}], && \underline{{r_2}_i} \in {\rm I\!R}^{d_{BERT}+d_{POS}}
\end{align}

Our GNN model generates graph embedding based on these concatenated features, which are then concatenated together to arrive at our final embeddings.

\begin{align}
\underline{{r_3}_i} = GNN(\underline{{r_2}_i}), && \underline{{r_3}_i} \in {\rm I\!R}^{d_{GNN}} \\
\underline{{r_4}_i} = [\underline{{r_2}_i}, \underline{{r_3}_i}], && \underline{{r_4}_i} \in {\rm I\!R}^{d}
\end{align}

$d_{GNN}$ refers to the output dimension for the last layer in our GNN module introduced in Section \ref{sec:dependency}. That is, if BiLSTM is opted, then $d_{GNN}$ refers to the size of the output dimension of the BiLSTM layer. If not, it refers to the output dimension for the second SAGEConv layer. Our final representations have a feature dimension size of $d = d_{BERT}+d_{POS}+d_{GNN}$.

Next, we run our combined embeddings through a linear layer to obtain predicted probabilities per token. $c$ refers to the number of classes to be predicted.
\begin{align}
\underline{{o}_i} = \underline{{r_4}_i}*W + \underline{b_i}, \nonumber \\ & W \in {\rm I\!R}^{d \times c}, & \underline{{o}_i}, \underline{{b}_i} \in {\rm I\!R}^{c}
\end{align}
Cross entropy loss was used during training to optimize model weights. In evaluation, the logits ran through a Viterbi decoder for adjustment. Finally, all logits ran through an argmax function to obtain the predicted class that had the highest probability. 

In our implementation, we set $d_{BERT}=768$, $d_{POS}=51$, $d_{GNN}=512$ and $c=5$.

\subsection{Replication Checklist}
\label{ssec:replication}
\begin{itemize}
    \item Hyperparameters: Our pretrained BERT models were initialized with the default configuration from Huggingface \citep{wolf-etal-2020-transformers}. To train our model, we used the Adam optimizer with $\beta 1 = 0.9$, $\beta 2 = 0.999$. Learning rate was set at $2e-05$ with linear decay. GPU train batch size was set as $4$. Maximum sequence length was $350$ tokens. For GNN, the graph hidden channel dimensions (i.e. output dimension of the first SAGEConv layer) was $1024$, the graph output dimension (i.e. output dimension of the second SAGEConv layer) was $512$, and the BiLSTM output dimension was also $512$. Probability for all dropout layers was $0.1$.
    \item Device: All experiments were ran on the NVIDIA A100-SXM4-40GB GPU.
    \item Time taken: For 3 folds over 10 epochs each, the Proposed model took us on average (over the 5 random seeds) $1 hour : 48 minutes : 28 seconds$ for \texttt{bert-base-cased} and $1 hour : 22 minutes : 24 seconds$ for \texttt{bert-large-cased} to train, validate and predict. For a single run over 10 epochs to generate our submission, the code took $28 minutes : 17 seconds$ and $20 minutes : 52 seconds$ to train and predict for the base and large models respectively.
\end{itemize}

\subsection{Qualitative Results}
In Table \ref{tab:qualitative}, we provide examples where the Proposed versus Baseline model predicts correctly when the other predicts wrongly. The predictions are obtained from the CV set of the \texttt{bert-large-cased} model with $seed=916$ and the first fold.

\begin{table*}[]
\resizebox{1.0\linewidth}{!}{
\begin{tabular}{p{8mm}p{75mm}p{75mm}p{8mm}}\hline

Index  & Baseline & Proposed & Right? \\\hline

0036 .000 11 & \hlorange{\texttt{<E>}Future sales agreements with suppliers increased during the period, and aggregate   contracted sales volumes are now 11.7m tonnes per annum\texttt{</E>}}, following   \hlgreen{\texttt{<C>}new European supply agreements.\texttt{</C>}}                   & \hlgreen{\texttt{<C>}Future sales agreements with suppliers increased during the period, and\texttt{</C>}} \hlorange{\texttt{<E>}aggregate contracted sales volumes are now 11.7m tonnes per annum\texttt{</E>}}, following new European supply agreements.                   & Base-line \\\hline

0270 .000 09  &  \hlorange{\texttt{<E>}
It comes with a \textsterling250 free overdraft and requires a \textsterling1,000 monthly deposit\texttt{</E>}} to \hlgreen{\texttt{<C>}avoid a \textsterling10 monthly fee.\texttt{</C>}} & \hlgreen{\texttt{<C>}It comes with a \textsterling250 free overdraft\texttt{</C>}} and requires a \textsterling1,000 monthly deposit to \hlorange{\texttt{<E>}avoid a \textsterling10 monthly fee.\texttt{</E>}} & Base-line \\\hline

0209 .000 33 & \hlgreen{\texttt{<C>}Fiserv believes that   this business combination makes sense from the complementary assets between the two companies, projecting higher revenue growth than\texttt{</C>}} \hlorange{\texttt{<E>}it would achieve on its own and costs savings of about \$900 million over five years.\texttt{</E>}} & \hlgreen{\texttt{<C>}Fiserv believes that   this business combination makes sense from the complementary assets between   the two companies\texttt{</C>}}, \hlorange{\texttt{<E>}projecting higher revenue growth than it would achieve on its own and costs savings of about \$900 million over five years.\texttt{</E>}} & Prop-osed \\\hline

0003 .000 19           &  \hlorange{\texttt{<E>}Additionally, the Congress provided \$125 million in the current fiscal year for sustainable landscapes programming\texttt{</E>}} to \hlgreen{\texttt{<C>}prevent forest loss.\texttt{</C>}} & \hlorange{\texttt{<E>}Additionally, the Congress provided \$125 million in the current fiscal year\texttt{</E>}} for \hlgreen{\texttt{<C>}sustainable landscapes programming to prevent forest loss.\texttt{</C>}} & Prop-osed \\\hline

\end{tabular}
}
\caption{Predicted Cause-Effect spans for CV set from $seed=916$ on first fold (i.e. $K0$). \emph{Notes.} Cause and Effect spans highlighted in green and orange respectively.}
\label{tab:qualitative}
\end{table*}

\end{document}